\documentclass{article}

\usepackage{microtype}
\usepackage{graphicx}
\usepackage{subfigure}
\usepackage{booktabs}
\usepackage{hyperref}

\usepackage[accepted]{icml2021}

\icmltitlerunning{COVID-Net Clinical ICU: ICU Admission Prediction for COVID-19 Patients}

\begin{document}

\twocolumn[
\icmltitle{COVID-Net Clinical ICU: Enhanced Prediction of ICU Admission for COVID-19 Patients via Explainability and Trust Quantification}
\icmlsetsymbol{equal}{*}

\begin{icmlauthorlist}
\icmlauthor{Audrey G.~Chung}{equal,DAI}
\icmlauthor{Mahmoud Famouri}{equal,DAI}
\icmlauthor{Andrew Hryniowski}{DAI,VIP,WAII}
\icmlauthor{Alexander Wong}{DAI,VIP,WAII}
\end{icmlauthorlist}

\icmlaffiliation{DAI}{DarwinAI Corp., Waterloo, Ontario, Canada }
\icmlaffiliation{VIP}{Vision and Image Processing Research Group, University of Waterloo, Waterloo, Ontario, Canada }
\icmlaffiliation{WAII}{Waterloo Artificial Intelligence Institute, Waterloo, Ontario, Canada }

\icmlcorrespondingauthor{Audrey G.~Chung}{audrey@darwinai.ca}
\icmlcorrespondingauthor{Mahmoud Famouri}{mahmoud@darwinai.ca}

\icmlkeywords{Machine Learning, ICML}

\vskip 0.3in
]
\printAffiliationsAndNotice{\icmlEqualContribution}

\begin{abstract}
The COVID-19 pandemic continues to have a devastating global impact, and has placed a tremendous burden on struggling healthcare systems around the world. Given the limited resources, accurate patient triaging and care planning is critical in the fight against COVID-19, and one crucial task within care planning is determining if a patient should be admitted to a hospital’s intensive care unit (ICU). Motivated by the need for transparent and trustworthy ICU admission clinical decision support, we introduce COVID-Net Clinical ICU, a neural network for ICU admission prediction based on patient clinical data. Driven by a transparent, trust-centric methodology, the proposed COVID-Net Clinical ICU was built using a clinical dataset from Hospital S\'{i}rio-Liban\^{e}s comprising of 1,925 COVID-19 patient records, and is able to predict when a COVID-19 positive patient would require ICU admission with an accuracy of 96.9\% to facilitate better care planning for hospitals amidst the on-going pandemic. We conducted system-level insight discovery using a quantitative explainability strategy to study the decision-making impact of different clinical features and gain actionable insights for enhancing predictive performance. We further leveraged a suite of trust quantification metrics to gain deeper insights into the trustworthiness of COVID-Net Clinical ICU. By digging deeper into when and why clinical predictive models makes certain decisions, we can uncover key factors in decision making for critical clinical decision support tasks such as ICU admission prediction and identify the situations under which clinical predictive models can be trusted for greater accountability.
\end{abstract}

\section{Introduction}
\label{intro}
The COVID-19 pandemic continues to have a devastating impact on the health and well-being of the global population, with far-reaching social and economic effects as shown by the World Health Organization~\cite{WHO2020}. In particular, COVID-19 has placed a tremendous burden on struggling healthcare systems around the world, depleting already scarce resources. A critical component of the clinical workflow in fighting COVID-19 is accurate triaging and care planning, which enables patient-centric personalized care while simultaneously reducing the load on hospitals by leveraging only the necessary resources for each patient. To that end, one crucial task within care planning is determining if a patient should be admitted to a hospital’s intensive care unit (ICU), especially given the ongoing shortage of available ICU space~\cite{Emanuel2020,Li2020,Tyrrell2021}.

With the goal of supporting clinical decisions, one promising avenue is to leverage machine learning to help predict ICU admissions~\cite{Li2020_2,Cheng2020,Zhao2020,Heo2021} by harnessing the wealth of clinical data being collected for each patient (e.g., demographic information, vital signs, blood results, etc.). However, a key challenge with building and using such predictive models is the difficulty understanding the rationale behind ICU admission predictions, the factors most critical to ICU admission, and under what circumstances a given predictive model is dependable and trustworthy. Motivated by the need for trustworthy clinical decision support and the potential for explainability to gain actionable insights into enhancing predictive performance, we introduce COVID-Net Clinical ICU\footnote{Available open source at: \textit{https://github.com/darwinai/covidnet\_clinical\_ICU}}, a neural network designed for ICU admission prediction based on clinical data built using a transparent, trust-centric methodology.

\begin{figure*}[t]
\begin{center}
\includegraphics[width=\linewidth]{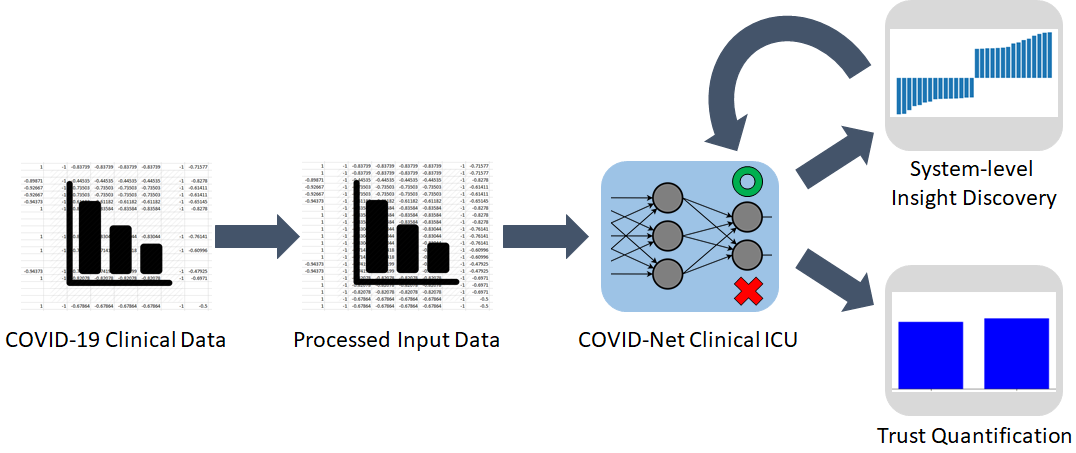}
\end{center}
\label{fig:algdesign}
\caption{Overview of transparent, trust-centric design methodology for COVID-Net Clinical ICU.}
\end{figure*}

\section{Data Preparation}
\label{data}
COVID-Net Clinical ICU is built using a clinical dataset from Hospital~\cite{SirioLiban2020}, comprising of 1,925 COVID-19 records from 385 patients and their associated demographic information (e.g., age and gender), information on previous diseases (e.g., hypertension, immunocompromised, etc.), blood results (e.g, platelets count, neutrophils count, etc.), and vital signs (e.g., body temperature, pulse rate, etc.) for a total of 228 clinical features. In this dataset, each patient record contains medical data at five different time cycles (i.e., 0-2, 2-4, 4-6, 6-12 and 12+ hours since hospital admission); as the patient can be admitted to the ICU at any time, only medical data prior to ICU admission was used to predict ICU admission and included in the training and testing sets. Patients who were admitted to the ICU at any time cycle were given a positive label, and patients with no record of ICU admission were given a negative label.

As typical of real-world data, the clinical dataset from Hospital S\'{i}rio-Liban\^{e}s contains samples with missing values. As it is impossible to examine all patients at all time cycles, many patient records have missing blood results and vital signs. This is especially true for patients with stable vital signs who are examined less frequently. In this study, we fill in the missing values with the latest available values from the previous time cycles of the same patient. After generating the ICU admission labels and filling in missing values, the dataset was split into 70\% training and 30\% testing.

\section{Neural Network Design Methodology}
\label{model}
\begin{table*}[h]
\centering
\caption{Summary of the initial and final COVID-Net Clinical ICU networks. The final network has a noticeably higher prediction accuracy while having 10\% fewer parameters. All the reported values are on the test set.}
\vspace{5pt}
\begin{tabular}{|c|c|c|c|c|c|}
\hline
    & \textbf{Architecture}   &   \textbf{No. of Parameters} &  \textbf{Sensitivity}     & \textbf{Specificity}   & \textbf{Accuracy} \\ \hline
Initial     &   228 - 220 - 100 - 5 - 1  & 125,203 & 91.3\% & 98.2\% & 95.7\% \\
Final    &   178 - 220 - 100 - 5 - 1  & 113,803 & 94.0\% & 98.5\% & 96.9\% \\ \hline
\end{tabular}
\label{tab_results}
\end{table*}

Based on the constructed training dataset, we built an initial neural network with a multi-layer fully-connected architecture. System-level insight discovery using a quantitative explainability strategy was then employed on the trained initial neural network to identify the clinical features exhibiting positive impact on the network's decision-making process. Leveraging these insights, a set of 178 clinical features was used to build the final COVID-Net Clinical ICU network. Figure~\ref{fig:algdesign} shows an overview of the design methodology used in this study. Both the initial neural network and the proposed COVID-Net Clinical ICU neural network were trained using the Adam optimizer with a binary cross entropy loss function for a total of 1,000 epochs. Learning weight decay policy with initial value of 0.001 has been used in the training process. All construction, training, and evaluation was conducted using TensorFlow Keras.

It can be seen from Table~\ref{tab_results} that the final COVID-Net Clinical ICU network is able to predict when a COVID-19 positive patient would require ICU admission with an accuracy of 96.9\% (noticeably higher than the initial neural network) while achieving lower architectural complexity at 10\% fewer parameters. This illustrates the importance of leveraging actionable insights gained from explainability for enhancing predictive performance and enabling better care planning for hospitals amidst the on-going pandemic. The explainability methodology for system-level insight discovery as well as the trust quantification methodology leveraged in this study are detailed below.

\begin{figure}[t]
\begin{center}
\includegraphics[width=\linewidth]{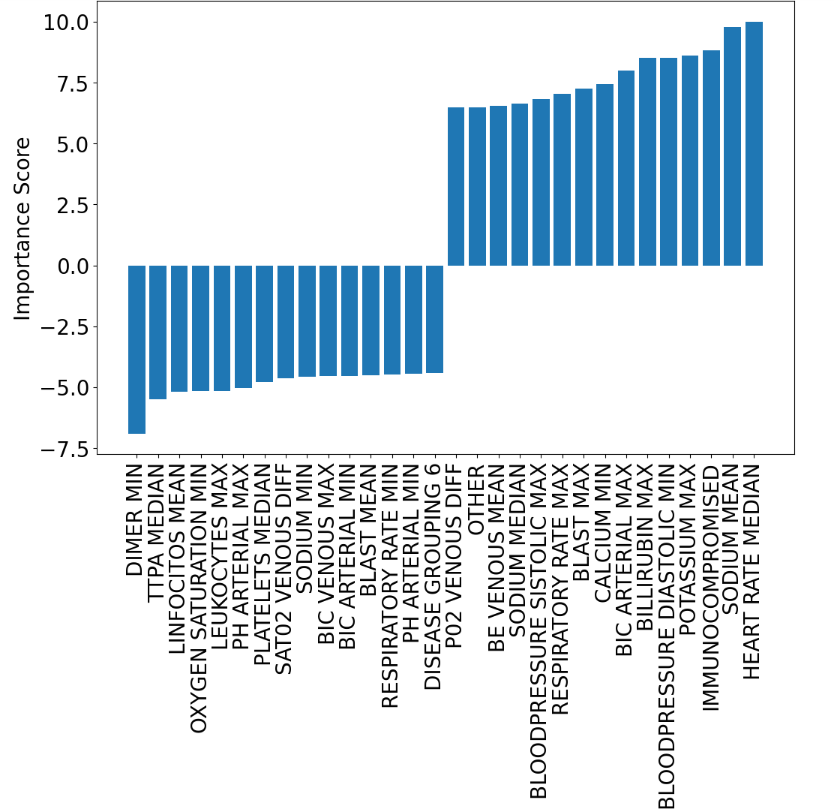}
\end{center}
\caption{The 15 most important and the 15 least important clinical factors for ICU admission for the initial neural network across the clinical dataset from Hospital S\'{i}rio-Liban\^{e}s.}
\label{fig:XAI_full_dataset}
\end{figure}

\subsection{System-level Insight Discovery via Explainability}
\label{xai}
To extract valuable insights into the decision-making process of the initial neural network, we conducted system-level insight discovery on COVID-Net Clinical ICU using a quantitative explainability strategy. In this study, we leverage the GSInquire method proposed by~\cite{Lin2019} as the explainability strategy of choice, which was shown to provide explainability insights that better reflect the decision-making process of neural networks compared to other state-of-the-art explainability methods. More specifically, we leveraged GSInquire to gain valuable actionable insights by determining the quantitative impact of the 228 available clinical features on the ICU admission prediction of the initial neural network across the training dataset. As a result, we were able to build the final COVID-Net Clinical ICU neural network with enhanced predictive performance compared to the initial neural network by leveraging only clinical features identified to have positive impact on the decision-making process. Furthermore, such a system-level insight discovery process enables greater transparency into the decision-making behaviour of the neural network, and enables insights into what may be useful for better supporting clinical decisions.

Figure~\ref{fig:XAI_full_dataset} shows the 15 most predictive (positive quantitative impact) and the 15 least predictive (negative quantitative impact) clinical factors used by the initial neural network for ICU admission. It can be observed that the most predictive factors for predicting whether a COVID-19 positive patient should be admitted to ICU are their median heart rate, mean blood sodium level, and if they are immunocompromised, as they have very high quantitative impact on the predictive performance of the network. Conversely, it can be seen that the minimum blood D-dimer level, median partial thromboplastin time (TTPA), and the average lymphocyte (linfocitos) count are the least predictive clinical factors. These interesting insights highlight the importance of taking a system-level insight discovery strategy using a quantitative explainability approach to better understand what factors are most important for informing clinical decisions pertaining to ICU admission for COVID-19 patients.

Based on the system-level insight discovery process, a total of 50 clinical features with negative quantitative impact were identified. Excluding these negatively impacting clinical features led to a set of 178 clinical features from the Hospital S\'{i}rio-Liban\^{e}s clinical dataset for building the final COVID-Net Clinical ICU network with enhanced predictive performance.

\subsection{Trust Quantification}
\label{trust}
\begin{figure*}[h]
\centering
\begin{tabular}{p{0.47\linewidth} p{0.47\linewidth}}
\includegraphics[width=\linewidth]{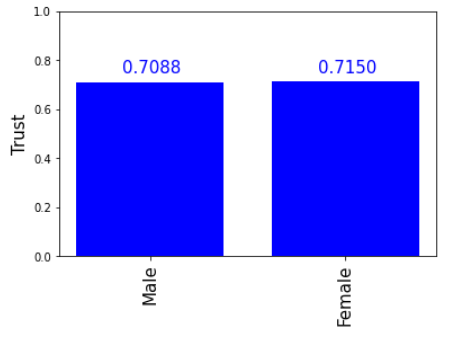} & \includegraphics[width=\linewidth]{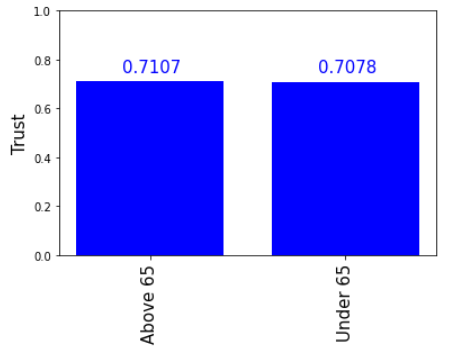} \\ \\
~~~~~~~~~~~~~~~~~~~~~~~~~~~(a) Trust spectrum for gender & ~~~~~~~~~~~~~~~~~~~~~~~~~~~(b) Trust spectrum for age \\
\end{tabular}
\caption{The demographic trust spectra for gender and age. The neural network generally behaves in a fair manner for both gender and age demographics. In particular, the network produces similarly trustworthy predictions for both male and female patients despite notably greater number of male patients compared to female patients in the clinical dataset.}
\label{fig:XAI_trust}
\end{figure*}

To gain deeper insights into the trustworthiness of the neural network, we evaluate the final COVID-Net Clinical ICU network at different levels of granularity using a suite of interpretable trust quantification metrics~\cite{Wong2020,Wong2020_2,Hryniowski2020}, as shown in Figure~\ref{fig:XAI_trust}. In particular, we take a closer look at the demographic trust spectrum to identify potential bias and gaps in fairness. While no model is perfect, understanding these gaps in fairness enables us to improve the overall performance and consistency of the model, as well as revealing when and where a neural network is dependable and providing trustworthy predictions.

It can be observed that the final COVID-Net Clinical ICU network generally behaves in a fair manner when making predictions across different demographics. The disparity in trustworthiness for patients over the age of 65 and patients aged 65 and under is minimal (i.e., 0.7107 vs. 0.7078) and, on the whole, the network is relatively fair when it comes to the trustworthiness of predictions made for both age demographic groups. More interesting, it can be observed that the neural network provides similarly trustworthy predictions for female and male patients (i.e., 0.7088 vs. 0.7150). This is particularly compelling given the fact that the neural network still behaves fairly for both male and female patients despite there being a notably higher number of male patients in the dataset compared to female patients (1,215 male vs. 710 female). This insight into the fairness of the final COVID-Net Clinical ICU network brings to light that the overall balance in the quantity of cases across demographic groups may not paint a complete picture in the resulting decision-making behaviour of neural networks that are built using the dataset. Furthermore, the trust quantification process can be a powerful tool for improved  trustworthiness and fairness if trust gaps are indeed identified.

\section{Conclusion}
\label{conclusion}
In this study, we presented COVID-Net Clinical ICU, a neural network for ICU admission prediction based on patient clinical data. Driven by a transparent, trust-centric methodology, the COVID-Net Clinical ICU network is able to predict when a COVID-19 positive patient would require ICU admission with a sensitivity of 94.0\%, a specificity of 98.5\%, and an overall accuracy of 96.9\%. We conducted system-level insight discovery on COVID-Net Clinical ICU via a quantitative explainability strategy to gain actionable insights for enhancing predictive performance, and leveraged a suite of trust quantification metrics to identify potential bias, gaps in fairness, and gain deeper insights into the trustworthiness of COVID-Net Clinical ICU. We hope that the public release of COVID-Net Clinical ICU can motivate and enable researchers, clinical scientists, and citizen scientists to accelerate progress in the field of AI for supporting the fight against the pandemic. Future work includes exploring the use of this transparent, trust-driven strategy when building neural networks for other important clinical decision support tasks such as mortality prediction, treatment recommendation, and outbreak prediction. Furthermore, it would also be interesting to explore decision-level insight discovery to further investigate and gain deeper insights into the decision-making process of COVID-Net Clinical ICU.

\bibliography{COVIDNet_Clinical_ICU}
\bibliographystyle{icml2021}

\end{document}